\documentclass[letterpaper, 10 pt, journal, twoside]{IEEEtran}  

\IEEEoverridecommandlockouts                              

\DeclareMathAlphabet{\mathcal}{OMS}{cmsy}{m}{n}

\usepackage{graphics}
\usepackage{epsfig}
\usepackage{mathptmx}
\usepackage{times}
\usepackage{amsmath}
\usepackage{amssymb}
\usepackage{xcolor}
\usepackage{xspace}
\usepackage[utf8]{inputenc}

\usepackage{enumitem}
\usepackage{cuted}
\usepackage{multirow}
\usepackage{float}
\usepackage{mathtools}
\usepackage{textcomp}

\usepackage{support-caption}
\usepackage{subcaption}
\usepackage[breaklinks=true,bookmarks=false,
             pdfauthor={Saini et al.},
             pdftitle={SmartMocap: Joint Estimation of Human and Camera Motion using Uncalibrated RGB Cameras},
             pdfsubject={SMartMocap}]{hyperref}
\DeclareMathSizes{10}{9}{6}{4}
\hypersetup{
    colorlinks=true,
    linkcolor=blue,
    filecolor=magenta,      
    urlcolor=blue,
}

\usepackage{soul}

\usepackage[font=footnotesize,labelfont=bf]{caption}
\title{SmartMocap: Joint Estimation of Human and Camera Motion using Uncalibrated RGB Cameras}

\author{Nitin Saini$^{1,2}$, Chun-Hao P. Huang$^{1}$, Michael J. Black$^{1}$, and Aamir Ahmad$^{2,1}$
\thanks{Manuscript received: December, 09, 2022; Revised January, 18, 2023; Accepted March, 13, 2023.}
\thanks{This paper was recommended for publication by Editor Gentiane Venture upon evaluation of the Associate Editor and Reviewers' comments.} 

\thanks{$^1$Max Planck Institute for Intelligent Systems, 72076 Tübingen, Germany. {\tt\footnotesize {firstname.lastname}@tuebingen.mpg.de}}%
\thanks{$^2$Institute of Flight Mechanics and Controls, University of Stuttgart, 70569 Stuttgart, Germany. {\tt\footnotesize {firstname.lastname}@ifr.uni-stuttgart.de}}%
\thanks{Digital Object Identifier (DOI): see top of this page.}
}

\newcommand{\humor}{\mbox{HuMoR}\xspace}

\usepackage{tabularx,booktabs}
\usepackage{multirow}

\begin{document}

\maketitle

\begin{abstract}
   Markerless human motion capture (mocap) from multiple RGB cameras is a widely studied problem. 
Existing methods either need calibrated cameras or calibrate them relative to a static camera, which acts as the reference frame for the mocap system. 
The calibration step has to be done \emph{a priori} for every capture session, which is a tedious process, and re-calibration is required whenever cameras are intentionally or accidentally moved. 
In this paper, we propose a mocap method which uses multiple static and moving extrinsically uncalibrated RGB cameras.  
The key components of our method are as follows. First, since the cameras and the subject can move freely, we select the ground plane as a common reference to represent both the body and the camera motions unlike existing methods which represent bodies in the camera coordinate system. Second, we learn a probability distribution of short human motion sequences ($\sim$1sec) relative to the ground plane and leverage it to disambiguate between the camera and human motion. Third, we use this distribution as a motion prior in a novel multi-stage optimization approach to fit the SMPL human body model and the camera poses to the human body keypoints on the images. 
Finally, we show that our method can work on a variety of datasets ranging from aerial cameras to smartphones. It also gives more accurate results compared to the state-of-the-art on the task of monocular human mocap with a static camera. A video demo and our code are available at \url{https://tinyurl.com/yeykrb67} and \url{https://tinyurl.com/2p9rme9y}.

\end{abstract}

\begin{IEEEkeywords}
  Gesture, Posture and Facial Expressions; Human Detection and Tracking; Deep Learning for Visual Perception
  \end{IEEEkeywords}

\section{Introduction}\label{sec:intro}

\setlength{\belowcaptionskip}{-20pt}
\setlength{\abovecaptionskip}{-5pt}
\begin{figure*}[ht!]
\begin{subfigure}{\textwidth}
  \centering
  \includegraphics[width=\linewidth]{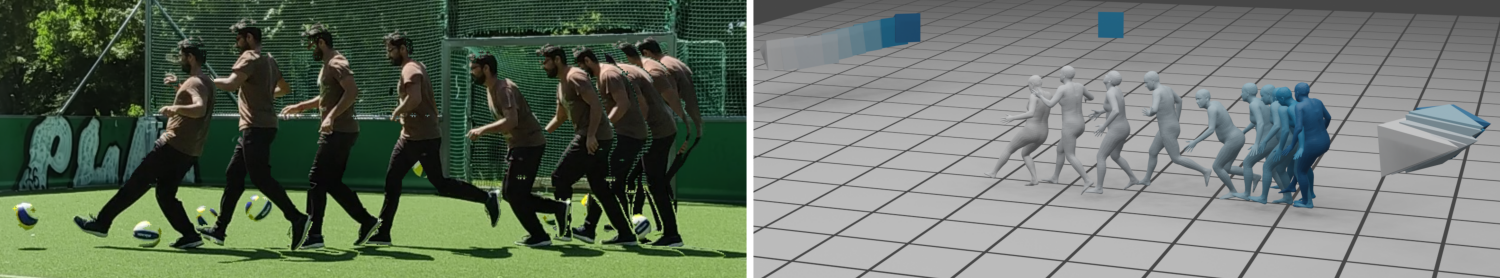}
  \label{fig:sfig1}
\end{subfigure}%
\caption{Multi-exposure image of a person playing football [left] and the reconstructed motion of the person and the cameras using our method [right].}
\label{fig:fig}
\end{figure*}
\setlength{\abovecaptionskip}{5pt}
\setlength{\belowcaptionskip}{-10pt}

\IEEEPARstart{M}{odern} markerless methods use RGB cameras to estimate human motion without the need for markers or sensors on the subject's body \cite{Nitin_ICCV_19}. They use sparse 2D keypoints to either fit a 3D body model or train a neural network to output the parameters of the body model. 
Existing monocular methods \cite{SMPL-X:2019} take images from a single static camera and estimate the motion of the person relative to it. Since, most applications need human motion relative to the world, the camera should be calibrated relative to the world. Another problem with this setup is that the subject's body parts can often get self-occluded. One solution is to make the camera freely moving such that it can optimize its view for motion estimation \cite{aircaprl}. 
However, a moving camera is much more difficult to calibrate relative to the world \cite{Nitin_ICCV_19}. Estimating the subject's global motion is also difficult using a single camera because the camera motion and the subject's motion cannot be disambiguated using only the sparse 2D keypoints. 
Multi-view methods employ multiple static cameras to handle self-occlusions. 
They calibrate the cameras relative to the world in a separate calibration step, which increases the preparation time. The cameras should remain static after the calibration step. In case they are moved intentionally or accidentally, the calibration has to be performed again, making the capture process highly inconvenient and time-consuming.

To address the aforementioned issues, in this paper, we present a system for outdoor human mocap using a set of RGB cameras, where some cameras are static while others are moving. This system is quick to set up, as users can place the cameras and immediately start the capture session. Any camera can be moved during the mocap session to get better visibility of the subject, and each camera, using our method, extrinsically calibrates itself relative to the world using only the sparse 2D keypoints of the human body.
Our system does not need a pre-calibration of the extrinsic parameters of the camera (pose of the camera in the 3D space). It, however, needs the camera intrinsics (related to the camera sensor and lens). Since these remain constant for any camera, the intrinsic calibration needs to be done only once and then can be used in multiple mocap sessions.

Our mocap method takes in the synchronized videos from multiple RGB cameras and estimates the camera poses and the subject's motion, defined as the trajectory of human poses (articulated and global), and shape in all the frames. All the estimates are relative to a global reference frame, which is the ground projection of the human root joint onto the ground plane in the first frame. The ground plane is defined as the XY plane. 
We learn a probability distribution of the human motion relative to a ground plane by training a variational autoencoder using a large human motion dataset (AMASS) \cite{AMASS:ICCV:2019}. The state-of-the-art human motion prior \cite{rempe2021humor} learns the distribution of pose transitions, which is defined as the difference between two consecutive poses. This is highly sensitive to noise in long-term motion generation/estimation. Contrarily, we learn the distribution of the trajectory of body joint positions and joint angles relative to the above-defined world frame. The length of each motion sequence is fixed as 25 frames at the rate of 30 frames per second. We use this probability distribution to fit the SMPL human body model \cite{SMPL:2015} to the sparse 2D keypoints in all the views. These 2D keypoints are obtained using the openpose 2D keypoint detector \cite{openpose}. While our motion prior encodes the distribution of human motion relative to the ground plane, the 2D keypoints contain information about the subject's articulated poses and the camera poses relative to the subject. In the optimization formulation, we directly optimize for the camera poses and the human poses in the world frame jointly, and condition the human motion using our learned probability distribution of human motions. This keeps the reconstructed poses aligned with the ground plane. However, this formulation is highly non-convex and susceptible to converging to a local minimum. Therefore, we first initialize the human position as the mean of human motion in the learned latent space. The articulated human pose and the camera poses are initialized using the estimates of a human pose regressor \cite{Kocabas_PARE_2021}. Since a motion sequence in our learned latent space is of a fixed length of 25 frames and starts from the origin, we take a multi-resolution optimization approach. After the initialization stage, we run the optimization stage, where we split the full sequence into chunks of 25 frames and run optimization on these chunks independently. This optimization treats the starting of each chunk as the origin. In the next stage (stitching stage), we stitch these chunks together such that the last frame of a sequence aligns with the first frame of the next sequence. In the final stage, we run the optimization again on this stitched sequence to get the final estimates. For longer sequences, stitching the full sequence can accumulate noise in the orientation estimates, which leads to poor initialization for the next optimization stage. To avoid this, we stitch together a smaller number of them, perform optimization, stitch, optimize and iterate. This way, we slowly increase the temporal resolution at each stitching stage and perform the alternate stitching and optimization until the final optimization for the full sequence.

In summary, we have the following novel contributions:
\begin{itemize}[left=-1pt]
    \item A human motion prior which encodes the global and articulated human motion relative to a global reference frame (ground plane).
    \item A multi-resolution optimization method for estimating camera poses along with the human motion and shape relative to the ground plane using single/multiple uncalibrated RGB cameras. 
\end{itemize}

\section{Related Work}
\label{sec:sota}
\subsection{Multi-view methods}
Most of the existing markerless human mocap methods utilize videos from multiple calibrated and time-synchronized cameras. They first detect 2D features (keypoints, silhouette etc.) on the image plane and then use the camera calibration parameters to either project them in 3D space \cite{iskakov2019learnable} or fit the 3D human body (model parameters, body joints etc.) to the 2D features \cite{muvs}. Since calibrating the cameras in a separate step is not always possible, \cite{takahashi} and \cite{huang3dv} utilize the human body to calibrate the static cameras. While \cite{takahashi} use simple human motion constraints such as constant velocity or acceleration of the human joints, \cite{huang3dv} learn a human motion prior and use it to fit the SMPL body model \cite{SMPL:2015} to the 2D keypoints. 
\cite{bachmann} use pan-tilt-zoom cameras and triangulate the human annotated 2D keypoints to reconstruct the 3d motion of people skiing and the pan, tilt and zoom of the cameras.
The static cameras cannot actively change the viewpoint for better mocap. Therefore, \cite{Elhayek2014} and \cite{Saini:IRL:2022} use moving cameras along with the static cameras as their mocap setup. However, all the above methods for uncalibrated cameras estimate the human motion relative to one static camera, which should be calibrated relative to the world such that the estimated human motion can be transformed to the world reference frame. Calibrating cameras is particularly hard if all the cameras are moving. 
\cite{hasler2009} use multiple handheld smartphone cameras and calibrates them relative to a static background using a structure-from-motion (SFM) method. However, such calibration is not reliable and only works for a static background with suitable texture.
\cite{Nitin_ICCV_19} use cameras mounted on custom-designed multiple micro-aerial vehicles and calibrates them using the onboard IMU and GPS sensors. However, such sensors are not available for ordinary RGB cameras.

\subsection{Monocular methods}\label{subsec:monocSOTA} 
Early monocular methods like \cite{guan2009}, fit a human body model \cite{Anguelov2005} to the 2D keypoints and body silhouette on the image plane.
Later ones use deep neural networks to directly regress the parameters of a human body model directly from the RGB image \cite{hmrKanazawa17}. All the above, estimate the human pose/motion relative to the camera or relative to some local coordinate system on the human body. Recent methods like \cite{rempe2021humor} try to estimate human motion in a world reference frame using a monocular video. \humor~\cite{rempe2021humor} learned the distribution of human motion transitions and used it to fit the SMPL model to the 2D keypoints on a monocular video from a static camera. However, encoding only the motion transitions is very sensitive to noisy observations (2D keypoints) which can lead to unrealistic human motion estimates. Methods like \cite{henning2022bodyslam} estimate human pose relative to the camera and use SLAM to track the camera poses that need textured background. Methods such as \cite{yuan2022glamr} train a regressor network which can output the trajectory of global pose given the articulated poses. This makes the final output more sensitive to the noisy articulated poses estimated in the previous stages. They do not utilize the fact that the articulated poses can also be improved using the global pose information, which we do. Additionally, they represent the global poses trajectory using relative local differences (similar to \cite{rempe2021humor}) which makes the future results sensitive to the noise in past poses.

\section{Approach}
\label{sec:methodology}
\subsection{Goal and preliminaries}\label{sec:prelim}
Given synchronized image sequences of length $T$ frames from $C$ cameras looking at a moving person, the goal is to estimate the camera motion, the person's shape, and the person's motion, which is defined as the trajectory of the person's poses (articulated and global). We use the SMPL human body model \cite{SMPL:2015} to represent the human poses. SMPL is parameterized by joint angles ($\theta \in \mathbb{R}^{63}$), body shape parameters ($\beta \in \mathbb{R}^{10}$), root orientation ($\phi \in \mathbb{R}^3$) and root position ($\tau \in \mathbb{R}^3$). We use $N=22$ body joints from SMPL, which includes 21 body joints and 1 root joint. We exclude the 2 hand joints from the original 24 joints in the SMPL model. Instead of representing the articulated pose as joint angles, we represent the subject's articulated pose at any time $t$ in the latent space of VPoser \cite{SMPL-X:2019} ($z \in \mathbb{R}^{32}$), which is a learned probability distribution of human poses. It is a variational autoencoder (VAE) with encoder ($\mathcal{V}_\mathcal{E}$) and decoder ($\mathcal{V}_\mathcal{D}$). The full human motion is then represented as $((\tau_1,\phi_1,z_1),...,(\tau_T,\phi_T,z_T),\beta)$.
The position and orientation of a camera $c$ at any time $t$ is represented as $p_{c,t} \in \mathbb{R}^{3}$ and, $r_{c,t} \in \mathbb{R}^{6}$ respectively. Unless explicitly stated, we use the 6D representation \cite{6drot} to represent the rotations in this paper. The camera motion for any camera $c$ is represented as $((r_{c,1},p_{c,1}),...,(r_{c,T},p_{c,T}))$.
Our human motion prior uses a different representation of the body pose. The body pose $x_t$ at any time $t$ is the orientation and position of each body joint relative to the world frame, i.e. $x_t \in \mathbb{R}^{22*(6+3)}$. The origin of the world frame is defined as the ground projection of the SMPL root joint in the first frame and the ground plane is defined as the XY plane. The motion prior encodes the fixed length of 25 consecutive poses, thus, the motion sequence for the motion prior is represented as $\mathbf{x} = (x_1,\dots,x_{25})$. We represent the estimated value of any parameter by putting a tilde over it, e.g. $\Tilde{\mathbf{x}}$ is the estimated value of $\mathbf{x}$.

\setlength{\belowcaptionskip}{-20pt}
\begin{figure*}
    \centering
    \includegraphics[width=\textwidth]{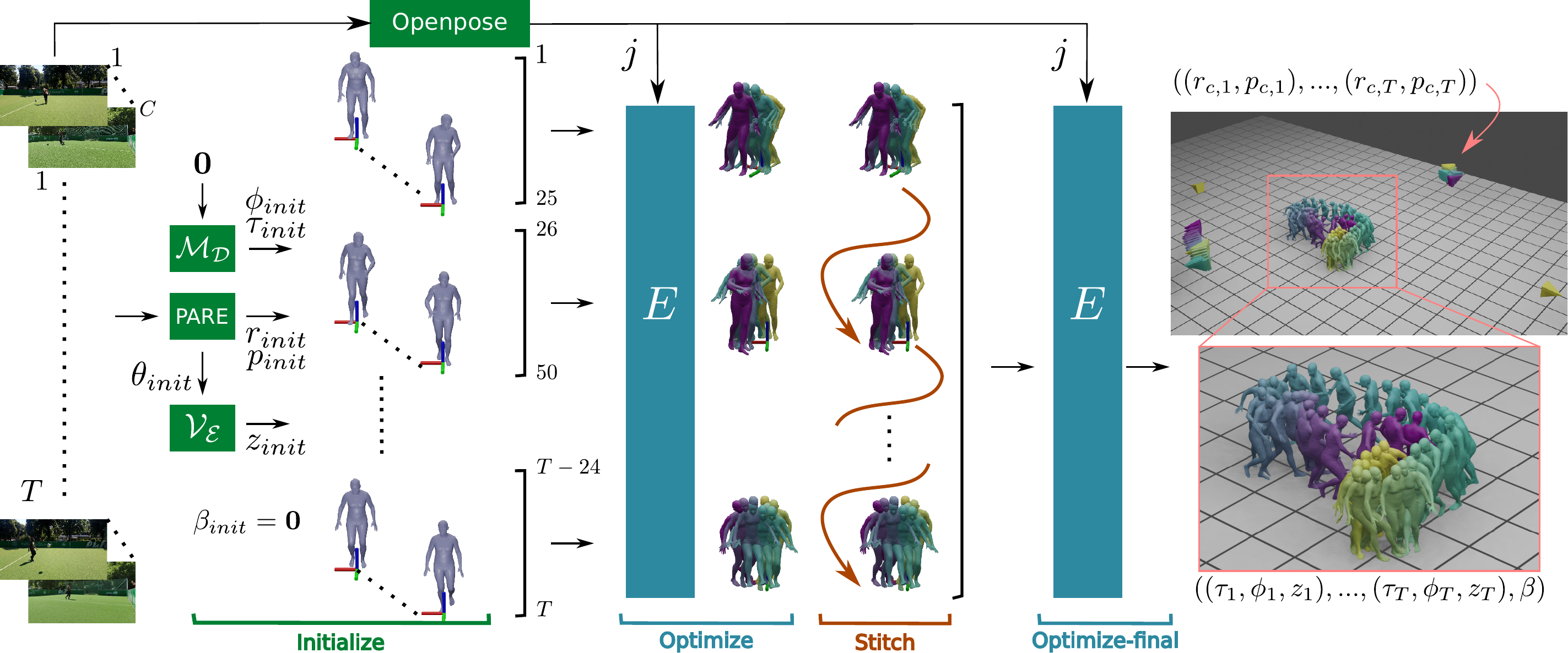}
    \caption{Our method takes in synchronized images of a moving person from multiple intrinsically uncalibrated cameras, and processes them in 4 steps 1) Initialize, 2) Optimize, 3) Stitch and 4) Optimize-final to give the motion of the person and the cameras in the world frame.}
    \label{fig:method}
\end{figure*}
\setlength{\belowcaptionskip}{-10pt}

\subsection{Human motion prior} \label{subsec:motionprior}
We use a VAE to learn a distribution in the latent space of human motion sequences of a fixed length. Each motion sequence consists of 25 consecutive human body poses at the rate of 30 frames per second. 
The forward-facing direction of the SMPL root joint in the first frame is aligned with the +Y axis of the origin.

\subsubsection{Training data}
We use AMASS dataset \cite{AMASS:ICCV:2019} to train our network. AMASS is a collection of multiple human mocap datasets, unifying them with the SMPL body representation. We follow the preprocessing steps in \humor~\cite{rempe2021humor} which removes the motion sequences where the person's feet are skating or sliding over the static ground plane. In such motions, the person doesn't interact with a stationary ground plane but with an object such as a treadmill or skates.
The preprocessing step also changes the frame rate to 30 FPS and gives out a total of 11893 motion sequences. We randomly select 25 consecutive frames from any of these sequences and canonicalize them such that the origin is the ground projection of the root joint at the first frame and the person's forward direction is aligned with the origin's +Y axis. Even though the shape of the subjects in AMASS varies, we follow \cite{rempe2021humor} and keep the body shape constant to the mean shape. This reduces the complexity of the model by ignoring the body shape variations at the expense of some possible artefacts such as foot-skating.

\subsubsection{Model architecture and training}
We use convolutional architecture for both the encoder and the decoder networks. The encoder ($\mathcal{M}_\mathcal{E}$) consists of a 1D convolutional (conv) layer at the input and 4 identical residual blocks. We modify the ResNet \cite{He_2016_CVPR} residual blocks to create these blocks. We replace the 2D convolutions with 1D convolutions. We further replace the ReLU units with the GELU units within the blocks. We also place a GELU unit after the input 1D conv layer and each of the residual blocks. The output dimension of the first conv layer is 1024. The input and output dimension of each residual block is 1024. Furthermore, two linear layers transform the output of the last residual block to the mean ($\mathbf{\mu} \in \mathbb{R}^{1024}$) and log of variance ($\log{(\mathbf{\sigma}^2)} \in \mathbb{R}^{1024}$) of the gaussian distribution in the latent space, from which the latent value ($m \in \mathbb{R}^{1024}$) is sampled using the reparametrization trick \cite{vae}. The decoder ($\mathcal{M}_\mathcal{D}$) architecture also consists of 4 consecutive residual blocks, similar to the ones in the encoder. The first residual block acts as the input layer, and there is a deconvolutional layer at the output of the decoder.

We train our motion VAE network using a combination of reconstruction ($\mathcal{L}_{rec}$) and KL divergence loss ($\mathcal{L}_{KL}$).
\begin{equation}
    \mathcal{L} = \mathcal{L}_{rec} + w_{kl}\mathcal{L}_{KL} ,
\end{equation}where
\begin{equation}
    \mathcal{L}_{rec} = ||\mathbf{x} - \mathbf{\Tilde{x}}||^2 ~~~ \mathrm{and} ~~~
    \mathcal{L}_{KL} = -\dfrac{0.5}{1024} \sum_i^{1024}(1+\log{(\sigma_i^2)}-\sigma_i^2-\mu_i^2).
\end{equation}

We employ a 20 epochs cyclic annealing scheme for the parameter $w_{kl}$ \cite{cyclicannealing}. Initially, the value of $w_{kl}$ starts at 0 and increases linearly with the training epochs. After 10 epochs, the value reaches 1 and stays constant for another 10 epochs. The value again drops to 0 and the cycle continues.

\subsection{Camera and human pose estimation}
First, we use openpose \cite{openpose} to detect 2D keypoints of the subject in each image. Then we use them in our method which consists of the following steps, 1) Initialize, 2) Optimize, 3) Stitch and 4) Optimize-final (see fig.~\ref{fig:method}).

\begin{itemize}[leftmargin=*, itemindent=9pt]
\item \textbf{Step 1: Initialize} We initialize the SMPL and camera poses for each frame using the results from PARE \cite{Kocabas_PARE_2021}. PARE gives the camera pose relative to the person and the person's articulated pose for each image. We take the mean of the articulated poses in all the views ($\theta_{init}$), project it to the VPoser latent space ($z_{init}$) and use it as the initial articulated pose of the subject. For the initialization of SMPL position ($\tau_{init}$) and orientation ($\phi_{init}$) relative to the ground plane, we use the decoded output of the mean value in the motion prior latent space. 
We use the initialized pose of the person to calculate the position ($r_{init}$) and orientation ($p_{init}$) of the cameras relative to the ground plane. Since PARE assumes an orthographic camera with a focal length of 5000, we use the method in \cite{kisso2020} to transform the SMPL position estimate in the actual camera. The SMPL shape ($\beta_{init}$) is initialized with a vector of zero values.

\item \textbf{Step 2: Optimize} We estimate the motion of the person in small intervals (25 frames). We split the full sequence into chunks of length 25 and run the optimization for each chunk independently in three phases to avoid local minimum, similar to \cite{Saini:IRL:2022}. 
In the first phase, we optimize the camera poses only. In the second phase, we optimize camera poses along with SMPL position and orientation. In the final phase, we optimize all the parameters, except the SMPL position and orientation in the first frame. The initial SMPL position and orientation in the first frame act as the pivot for all the optimizing parameters.

\item \textbf{Step 3: Stitch} We stitch together the estimated motions. Since the origin for each chunk is defined as the ground projection of the root joint, we stitch consecutive sequences together such that the root ground projection of the last frame of a chunk is aligned with the first frame of the next chunk. For very long sequences, we stitch together fewer sequences, optimize, stitch and repeat until all the sequences are stitched.

\item \textbf{Step 4: Optimize-final} In the final optimization step (optimize-final), we again optimize all the parameters for the fully stitched sequences in three phases, the same as in the previous optimization stage. This step is the final optimization step if all the sequences are stitched. For very long sequences, we stitch together fewer chunks instead of all. Then we optimize and repeat the stitching and optimization cycle until the whole sequence is done.  
\end{itemize}
 In all the optimization stages, we minimize the same loss function, which is a weighted combination of multiple loss terms. It is given as
\begin{equation}
\begin{split}
    E &= w_{2D}E_{2D} + w_{m}E_{m} + w_{3DS}E_{3DS} + w_{COS}E_{COS} + w_{CPS}E_{CPS} \\ 
    &+ w_{\beta}E_{\beta} + w_{z}E_{z} + w_{GP}E_{HGP} + w_{CGP}E_{CGP}.
\end{split}
\end{equation}

The component $E_{2D}$ is the 2D reprojection loss given as
\begin{equation}\label{eq:loss_2d}
\begin{split}
    E_{2D} = \dfrac{1}{NT}\sum_{n,c,t}w_n ||  \Pi(r_{c,t},p_{c,t}, \mathcal{J}_n(\mathcal{V}_\mathcal{D}(z_t),\tau_t,\phi_t,\beta)) - j_{c,t}^n ||^2,
\end{split}
\end{equation}where, $\mathcal{J}$ is the SMPL 3D joint regressor function \cite{SMPL:2015}, $\mathcal{V}$ is the VPoser decoder \cite{SMPL-X:2019}, $\Pi$ is the camera projection function, $j_{c,t}^n$ is the 2D keypoint corresponding to the joint $n$ in camera $c$ at time instant $t$, and $w_n$ is the confidence score given by the 2D detector for the joint $n$. The loss component $E_m$ is the motion prior loss given as 
\begin{equation}\label{eq:motion_reg}
    E_m = \sum_t^{T-25}||\mathcal{M}_{\mathcal{E}_{\mu}}(\mathcal{V}_\mathcal{D}(z_{t:t+25}),\tau_{t:t+25},\phi_{t:t+25},\beta)||^2,
\end{equation}where $\mathcal{M}_{\mathcal{E}_{\mu}}$ is the $\mu$ part of the motion prior encoder. The loss component $E_{3DS}$ is a temporal smoothing term for the 3D joint positions. It is given as
\begin{equation}
E_{3DS} = \sum_t||\mathcal{J}(\mathcal{V}_\mathcal{D}(z_t),\tau_t,\phi_t,\beta) -  \mathcal{J}(\mathcal{V}_\mathcal{D}(z_{t-1}),\tau_{t-1},\phi_{t-1},\beta)||^2.
\end{equation}

$E_{COS}$ and $E_{CPS}$ are the camera motion smoothing terms. Following \cite{Saini:IRL:2022}, we use L2 loss on the positions and the 6D representation of the camera orientations, given as \begin{equation}\label{eq:cam_smooth}
    E_{COS} = \dfrac{1}{CT}\sum_{c,t}||r_{c,t} - r_{c,t-1}||^2 ~~;~~ E_{CPS} = \dfrac{1}{CT}\sum_{c,t}||p_{c,t} - p_{c,t-1}||^2.
\end{equation}

$E_{\beta}$ and $E_{z}$ are the SMPL shape and VPoser regularization terms \cite{Saini:IRL:2022}, given as
\begin{equation}
    E_{\beta} = ||\beta||^2  ~~~ \textrm{and} ~~~ E_{z} = ||z||^2.
\end{equation}

$E_{HGP}$ and $E_{CGP}$ are the ground penetration terms for both the human and the cameras. These terms avoid the scenarios where the cameras or the human goes below the ground plane. These are given as
\begin{equation}
    E_{HGP} = \dfrac{1}{T}\sum_t max(0,\mathcal{J}^z(\mathcal{V}_\mathcal{D}(z_t),\tau_t,\phi_t,\beta_t)) ~~~ \textrm{and}
\end{equation}
\begin{equation}
    E_{CGP} = \dfrac{1}{CT}\sum_{c,t}max(0,p^z_{c,t}),
\end{equation}where, $\mathcal{J}^z$ and $p^z$ are the vertical ($z$) component of the 3D joint positions and the camera positions, respectively. In all the above equations, $T$ is replaced with 25 in the \textbf{Step 2}.

\section{Experiments and Results}
\label{sec:exp_results}

\subsection{Datasets}\label{subsec:datasets}
We evaluate our method using a sequence taken from each of the following datasets
\par 1) RICH dataset \cite{rich:Huang:CVPR:2022}: it is collected using 7 static and one moving IOI cameras. It has the ground truth poses of the person and the camera poses for the static cameras. The GT poses of the moving camera are not available. 
\par 2) AirPose real-world dataset \cite{Saini:IRL:2022}: It is collected using RGB cameras mounted on two DJI unmanned micro aerial vehicles (UAVs). One UAV is kept hovering and other is encircling the subject while looking at him.
\par
\begin{figure}
    \centering
    \includegraphics[width=\linewidth, trim={0cm 1.5cm 0 1.5cm},clip]{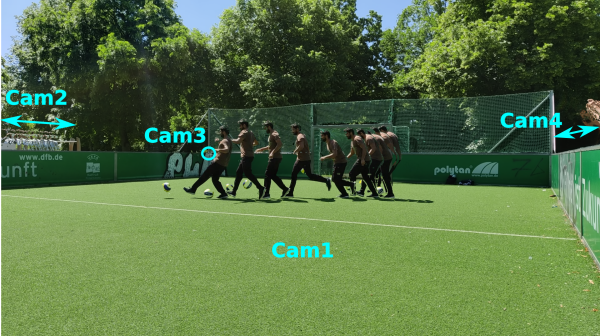}
    \caption{Camera setup used to collect our smartphone dataset.}
    \label{fig:volley_data_setup}
\end{figure}

 3) Our smartphone dataset: it is collected using 4 smartphone cameras, where two of the cameras are static while the other two are moving. 
The data is collected when the subject is 
playing with a football in a small 
field. The camera setup can be seen in Fig.~\ref{fig:volley_data_setup}. We show the rest of the three cameras and the subject in the frame of the camera \textit{Cam1}. \textit{Cam1} and \textit{Cam3} are static, and \textit{Cam2} and \textit{Cam4} are moved along the boundary walls of the field. We used OpenCamera app \cite{opencam} to collect the video data on the smartphones.  
Each smartphone records the videos at 30 FPS.
Since the frame rates are the same and constant, we manually synchronize the videos by synchronizing just one frame. Similar to AirPose dataset, the smartphone dataset also does not contain any GT.

\setlength{\belowcaptionskip}{-15pt}
\setlength{\belowcaptionskip}{-10pt}

\subsection{Metrics}\label{subsec:metrics}
We use the following metrics to quantitatively evaluate the reconstruction by our method on the RICH dataset.
1) \textit{Mean camera position error (MCPE):} This is the mean value of the distance between the estimated and the GT position of all the cameras. It is given as
\begin{equation}
    MCPE = \dfrac{1}{C}\sum_c ||p_c - \Tilde{p_c}||.
\end{equation}

2) \textit{Mean camera orientation error (MCOE):} This metric is the mean geodesic distance between the estimated camera orientation and the GT on the 3D manifold of rotation matrices \cite{so3dist}. It is given as
\begin{equation}
    MCOE = \dfrac{1}{C}\sum_c\arccos{(0.5*(Tr(\Tilde{R_c}R_c^{\top})-1))},
\end{equation}where $R_c$ is the camera orientation matrix.

3) \textit{Mean position error (MPE):} This is the mean value of the distance between the estimated SMPL root position parameter and its GT value provided by the RICH dataset.
\begin{equation}
    MPE = \dfrac{1}{T}\sum_t ||\tau_t - \Tilde{\tau_t}||.
\end{equation}

4) \textit{Mean orientation error (MOE):} This metric is the mean geodesic distance between the estimated SMPL root orientation and the GT on the 3D manifold of rotation matrices \cite{so3dist}. It is given as
\begin{equation}
    MOE = \dfrac{1}{T}\sum_t\arccos{(0.5*(Tr(\Tilde{R_{\phi_t}}R_{\phi_t}^{\top})-1))},
\end{equation}where $R_{\phi_t}$ is the SMPL root orientation matrix at time $t$.

5) \textit{Root-aligned mean per-joint position error (RA-MPJPE):} This metric is to quantitatively evaluate the articulated pose estimate \cite{Nitin_ICCV_19}. It is the distance between the estimated 3D joints and their corresponding values when the SMPL position, orientation of the root joint and shape are aligned with their corresponding GT. For alignment, all these three parameters are set to zero. The error is given as
\begin{equation}
    RA{\text -}MPJPE = \dfrac{1}{NT}\sum ||\mathcal{J}_n(\theta_t) - \mathcal{J}_n(\Tilde{\theta_t})||.
\end{equation}

6) \textit{Mean per-vertex position error (MPVPE): } This metric is to evaluate the shape estimate relative to the GT. We do the SMPL forward pass using only the estimated and GT shape parameters and then calculate the mean distance between the corresponding vertices as
\begin{equation}
    MPVPE = \dfrac{1}{V}\sum ||\mathcal{S}_v(\beta) - \mathcal{S}_v(\Tilde{\beta})||,
\end{equation}
where $\mathcal{S}$ is the SMPL vertices regressor function \cite{SMPL:2015} and V is the number of vertices in the SMPL model.

\subsection{Results and discussion}
\subsubsection{RICH dataset}
\begin{figure}[t]
    \centering
    \includegraphics[width=\linewidth]{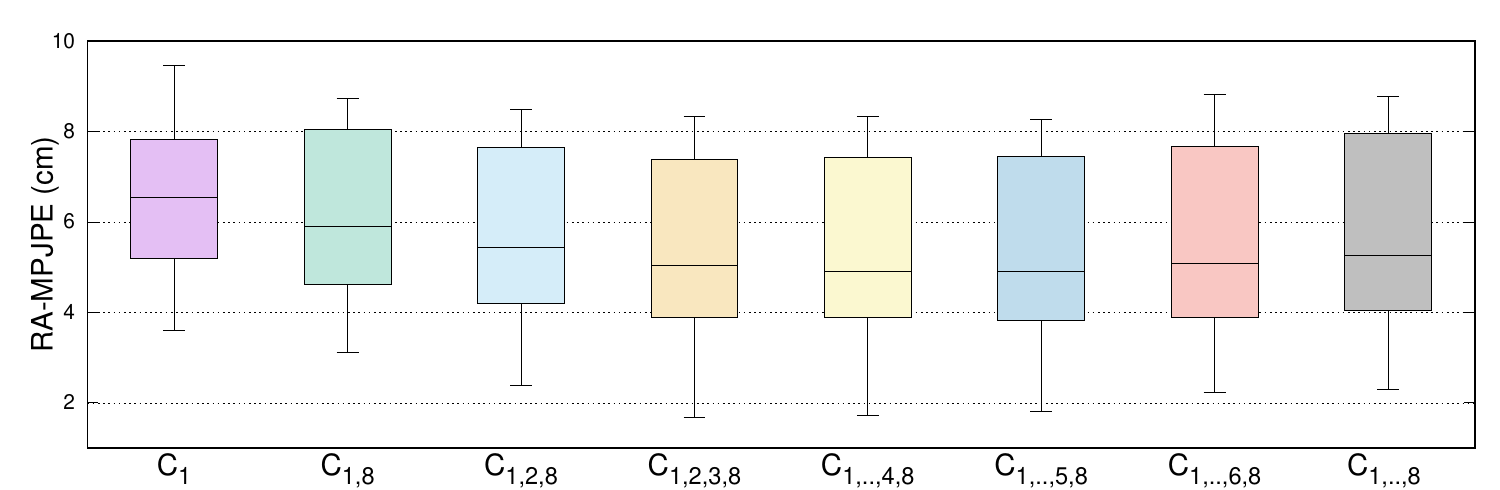}
    \caption{RA-MPJPE of our method using different camera configurations.}
    \label{fig:mpjpe_boxplots}
\end{figure}

We show the evaluation metrics and their corresponding standard deviation values of our method on RICH in Table \ref{tab:ablation} and \ref{tab:results}. The results of our method using all 8 cameras are shown in the last row of Table \ref{tab:ablation} (C$_{1,..,8}$). We also compare the performance of our method with multiple camera configurations. In the first row, we show the results when only the first camera is used (C$_1$). Next, we add camera 8, which is moving, and the results are shown in the second row (C$_{1,8}$). We keep adding static cameras one at a time and show the results in further rows.
Adding an extra view, only gives information about the articulated pose of the subject and the relative poses of the camera and the subject. 
This is why we don't see improvement in the global pose estimates of the person or in the camera poses with addition of more views. However, adding more views helps in handling occlusions better and we see in Fig.~\ref{fig:mpjpe_boxplots} the RA-MPJPE improves with more camera views but gets saturated after 4 views. This shows that 4 views are sufficient to resolve any uncertainty in the person's articulated pose due to occlusions, and adding more views doesn't provide any additional information.

Note that we do not optimize the SMPL position and orientation in the first frame, as it acts as the pivot and bring the person conforming to the first frame using (\ref{eq:motion_reg}) and cameras using (\ref{eq:loss_2d}). Therefore, the estimate in the first frame is noisy and because of (\ref{eq:cam_smooth}), a few initial frames become noisy. Hence, we ignore the first 10 frames for evaluation.

\setlength{\belowcaptionskip}{-5pt}
\begin{table}[t]\scalebox{0.75}{
\begin{tabular}{|l|l|l|l|l|l|l|l|}
\hline
Cameras       & \begin{tabular}[c]{@{}l@{}}MCPE\\ (cm)\end{tabular} & \begin{tabular}[c]{@{}l@{}}MCOE\\ (rad)\end{tabular} & \begin{tabular}[c]{@{}l@{}}MPE\\ (cm)\end{tabular} & \begin{tabular}[c]{@{}l@{}}MOE\\ (rad)\end{tabular} & \begin{tabular}[c]{@{}l@{}}RA-MPJPE\\ (cm)\end{tabular} & \begin{tabular}[c]{@{}l@{}}MPVPE\\ (cm)\end{tabular} \\ \hline
C$_1$           & 72.68                                                & 0.20                                                 & 10.89 $\pm$ 7.31                                               & 0.27 $\pm$ 0.10                                                & 6.60 $\pm$ 4.92                                                            & 3.07 $\pm$ 1.45                                                    \\ \hline
C$_{1,8}$       & 55.85                                                & 0.14                                                 & 7.61 $\pm$ 4.14                                               & 0.22 $\pm$ 0.06                                                & 6.24 $\pm$ 5.02                                                            & 3.13 $\pm$ 1.50                                                       \\ \hline
C$_{1,2,8}$    & 90.46                                                & 0.17                                                 & 12.72 $\pm$ 2.93                                               & 0.23 $\pm$ 0.07                                                & 5.97 $\pm$ 4.91                                                            & 3.00 $\pm$ 1.38                                                      \\ \hline
C$_{1,2,3,8}$   & 89.68                                                & 0.14                                                 & 11.67 $\pm$ 2.59                                               & 0.20 $\pm$ 0.07                                                & 5.73 $\pm$ 4.80                                                            & 3.02 $\pm$ 1.38                                                      \\ \hline
C$_{1,..,4,8}$ & 95.03                                                & 0.16                                                 & 12.39 $\pm$ 2.69                                               & 0.22 $\pm$ 0.07                                                & 5.68 $\pm$ 4.68                                                            & 2.91 $\pm$ 1.30                                                      \\ \hline
C$_{1,..,5,8}$ & 93.74                                                & 0.16                                                 & 12.49 $\pm$ 2.74                                               & 0.20 $\pm$ 0.07                                                & 5.66 $\pm$ 4.65                                                            & 2.91 $\pm$ 1.32                                                      \\ \hline
C$_{1,..,6,8}$ & 92.43                                                & 0.17                                                 & 13.42 $\pm$ 2.85                                               & 0.20 $\pm$ 0.06                                                & 5.87 $\pm$ 4.77                                                            & 2.31 $\pm$ 0.92                                                      \\ \hline
C$_{1,..,8}$   & 88.13                                                & 0.18                                                 & 13.69 $\pm$ 3.05                                               & 0.19 $\pm$ 0.06                                                & 6.03 $\pm$ 4.94                                                            & 1.86 $\pm$ 0.79                                                     \\ \hline
\end{tabular}}
\caption{Evaluation of our method using multiple camera configurations. Camera no. 1-7 are static in the RICH dataset with available GT, and camera no. 8 is moving but GT is not available. Hence, the MCPE and MCOE metrics for rows 2-9 do not include camera 8.}
\label{tab:ablation}
\end{table}
\setlength{\belowcaptionskip}{-10pt}

\setlength{\belowcaptionskip}{-20pt}

\begin{table}[]\scalebox{0.75}{
\begin{tabular}{|l|l|l|l|l|l|l|l|}
\hline
Camera                                                                           & \begin{tabular}[c]{@{}l@{}}MCPE\\ (cm)\end{tabular} & \begin{tabular}[c]{@{}l@{}}MCOE\\ (rad)\end{tabular} & \begin{tabular}[c]{@{}l@{}}MPE\\ (cm)\end{tabular} & \begin{tabular}[c]{@{}l@{}}MOE\\ (rad)\end{tabular} & \begin{tabular}[c]{@{}l@{}}RA-MPJPE\\ (cm)\end{tabular} & \begin{tabular}[c]{@{}l@{}}MPVPE\\ (cm)\end{tabular} \\ \hline
\begin{tabular}[c]{@{}l@{}}GLAMR \\ \cite{yuan2022glamr}\end{tabular}                                                                            & 250.93                                              & 0.27                                               & 24.07 $\pm$ 4.96                                               & 0.48 $\pm$ 0.28                                               & 9.66 $\pm$ 9.51                                                           & 2.84 $\pm$ 1.12                                                         \\ \hline
\begin{tabular}[c]{@{}l@{}}HuMor \\ \cite{rempe2021humor}\end{tabular}                                                                            & 90.09                                                & \textbf{0.17}                                               & 30.32 $\pm$ 8.12                                               & 0.48 $\pm$ 0.39                                               & 10.82 $\pm$ 8.14                                                           & 4.2 $\pm$ 2.09                                                         \\ \hline
\begin{tabular}[c]{@{}l@{}}C$_1$ \\ (ours)\end{tabular}                           & \textbf{72.68}                                       & 0.20                                       & \textbf{10.89} $\pm$ \textbf{7.31}                                               & 0.27 $\pm$ 0.10                                                & 6.6 $\pm$ 4.92                                                            & 3.07 $\pm$ 1.45                                                                              \\ \hline
\begin{tabular}[c]{@{}l@{}}C$_{1,..,8}$ \\ (ours)\end{tabular} & 88.13                                                & 0.18                                                & 13.69 $\pm$ 3.05                                      & \textbf{0.19} $\pm$ \textbf{0.06}                                       & \textbf{6.03} $\pm$ \textbf{4.94}                                                   & \textbf{1.86} $\pm$ \textbf{0.79}                                                                   \\ \hline
\end{tabular}}
\caption{Comparison of our method (C$_{1,..,8}$), monocular version of our method (C$_1$) and state-of-the-art monocular methods HuMor \cite{rempe2021humor} and GLAMR \cite{yuan2022glamr} on RICH dataset.}
\label{tab:results}
\end{table}

\setlength{\belowcaptionskip}{-10pt}

\begin{figure*}[h!]
    \centering
    \includegraphics[width=\textwidth]{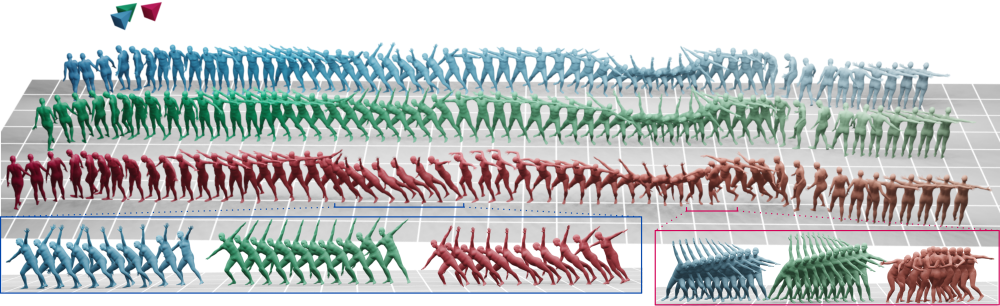}
    \caption{Comparison: our method (blue), HuMoR \cite{rempe2021humor} (red) and GT (green). For clarity, the human poses are shifted sideways using time-dependent offsets.}
    \label{fig:humor_comp}
\end{figure*}

\begin{figure*}[!htbp]
    \centering
    \includegraphics[width=\textwidth]{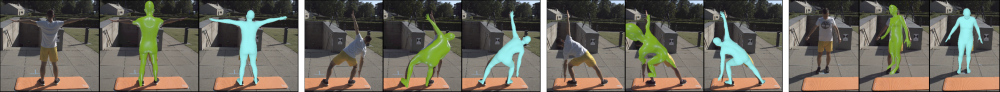}
    \caption{Comparison of GLAMR \cite{yuan2022glamr} and our method using a single camera on the RICH dataset. We show images at 4 time instants, each containing the original image, the GLAMR results (green) and results from our method (cyan) using a single camera.}
    \label{fig:glam_comparison}
\end{figure*}

In table \ref{tab:results}, we compare the monocular and the multi-view version of our method (row 3 and 4) with the reference methods GLAMR \cite{yuan2022glamr} (row 1) and HuMor \cite{rempe2021humor} (row 2), which are the state-of-the-art monocular human pose and shape estimation methods. We see that our method significantly outperforms these methods. Both HuMor and GLAMR uses a motion prior which encodes human motion transitions instead of absolute motions. As we discussed in Sec.~\ref{subsec:monocSOTA}, reconstructing the motion from the space of motion transitions is very sensitive to noise and can lead to spurious results. In Fig.~\ref{fig:humor_comp}, we show the qualitative results of our method and the reference method and compare them with the GT. The 3D reconstruction of the human and the cameras relative to the ground plane are shown for our method (blue), the GT (green) and the reference method \cite{rempe2021humor} (red). For a clearer illustration, we render each pose by adding a time-dependent offset to the position estimate at that time. Camera pose estimates are unchanged for the rendering. We can see that our estimates are very close to the GT, while the reference method estimates are quite inaccurate. For example, in the left inset box, we see that both the feet are on the same side, giving a physically implausible global pose estimate. In the right inset box, the person's body suddenly rotates more than $90^{\circ}$, again resulting in a physically implausible motion. In Fig.~\ref{fig:glam_comparison}, we do a qualitative comparison of our method with GLAMR \cite{yuan2022glamr} by showing the resulting mesh overlaid on the original image. While the results from our method are nearly perfectly aligned with the person in the image, the overlaid GLAMR results does not match the person. This is due to the errors in both the estimated person's poses and the estimated camera pose.

\setlength{\belowcaptionskip}{-15pt}
\setlength{\belowcaptionskip}{-10pt}

\setlength{\belowcaptionskip}{-20pt}

\begin{figure*}[!htbp]
    \centering
    \includegraphics[width=\textwidth]{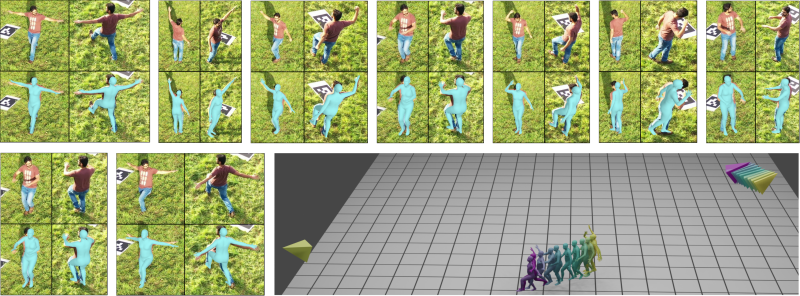}
    \caption{Results on the Airpose real-world dataset. The result at each time instant is shown using an image grid of 2×2. The top row shows the cropped region of the actual image, while the bottom row shows the estimated mesh overlaid on top of it. Each column corresponds to each camera view. The bottom-right image shows the full 3D reconstruction of the subject's poses, shape, and the camera poses.}
    \label{fig:copenet_res}
\end{figure*}

\setlength{\belowcaptionskip}{-20pt}

\begin{figure*}[!htbp]
    \centering
    \includegraphics[width=\textwidth]{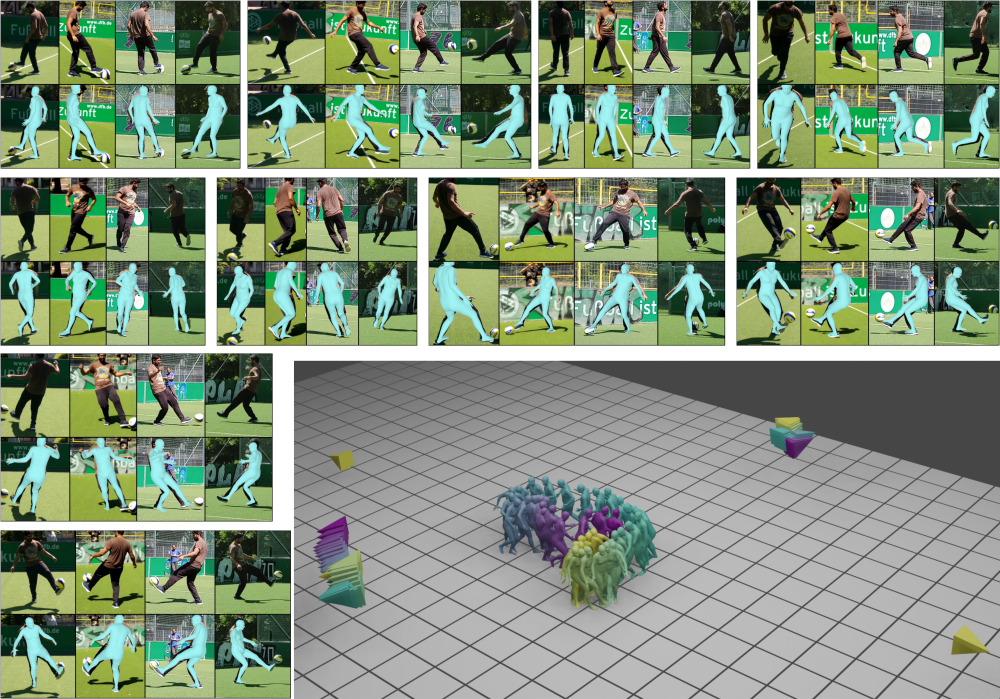}
    \caption{Results on our smartphone dataset. The result at each time instant is shown using an image grid of 2×4 (check Fig.~\ref{fig:copenet_res} caption for details).}
    \label{fig:smartphone_res}
\end{figure*}

\subsubsection{Airpose dataset}
In Fig.~\ref{fig:copenet_res}, we show qualitative results of our method on the Airpose real-world dataset. We show the cropped version of the original images of the subject, along with the same image with the estimated mesh overlaid on top. Two adjacent columns are the two views at the same time instant. The results show that our method can reconstruct the diverse poses captured from an aerial view. In the bottom-right corner, we show the 3D reconstruction of the subject's poses, shape, and the camera poses for a sub-sequence. The color gradient from yellow to violet is used to show the time transition. We observe that the subject's reconstructed body is not touching the ground, but lies a bit above the ground plane. This is because the actual terrain is not a plane, but a sloped hilly terrain. Even though the motion prior is trained on the motion sequences performed on a plane surface, our method can still recover the global motion on terrain with a small slope.

\subsubsection{Smartphone dataset}
We also show the qualitative results of our method on our smartphone dataset. We show the cropped images, the overlaid estimated mesh and the 3D reconstruction of the subject and the cameras in Fig.~\ref{fig:smartphone_res}. Our method accurately reconstructed the subject's pose playing football and the camera motion along the wall of the playing arena. We see that the overlays are near-perfectly aligned with the images, showing the accurate reconstruction of the relative pose of the camera and the person. The complete reconstruction is shown in the bottom-right image, and we see that the subject's motion and the camera motion are temporally and spatially coherent.

\section{Limitations}
\label{sec:limitations}
Our method assumes a planar ground surface and human motions which do not involve moving ground (e.g. a treadmill) or sliding motions (e.g. skating, skiing, etc.). However, it can be extended for non-planar ground surfaces by encoding the surface, articulated poses and global poses together in a prior. A major limitation in training such a prior network is the unavailability of human mocap data where the ground surface is also captured. Moreover, most existing datasets are collected with a human subject moving on a planar ground surface.

\section{Conclusions and Future Work}
\label{sec:conclusion}
In this paper, we presented a method to reconstruct the 3D human poses, shape, and camera poses relative to a global coordinate frame using synchronized RGB videos from single/multiple extrinsically uncalibrated cameras. We use the ground plane as the reference coordinate system and train a human motion prior using a large amount of human mocap data. We use the latent space of this motion prior to fit the SMPL body model to the 2D keypoints on all the views simultaneously. We show our results on two existing dataset and one new dataset that we collect using smartphones. We show that our method reconstructed the human poses, shape, and camera poses on all the three datasets. We showed the quantitative results on the RICH dataset, demonstrating that our method achieves more accurate results compared to a state-of-the-art method on the task of monocular human motion reconstruction. We also analyzed the effects of multiple views on our method's performance. We show that our method works for diverse types of camera views by showing qualitative results on all the three datasets. The accurate reconstruction by our method on the smartphone data is evidence of the ease of use of our method.  Our future work includes usage of synthetic data with physics to generate training dataset.

\begin{normalsize}
\textbf{Disclosure}: Michael J.\ Black has received research gift funds from Intel, Nvidia, Adobe, Facebook, and Amazon. While he is a part-time employee of Amazon, his research was performed solely at, and funded solely by, MPI. His has financial interests in Amazon and Meshcapade GmbH.
\end{normalsize}

{
	\bibliographystyle{IEEEtran}
	\bibliography{SmartMocap}
}

\end{document}